\pdfoutput=1
 
\documentclass[11pt]{article}

\usepackage[backref=page]{hyperref}

\usepackage{acl}

\usepackage{xcolor}
\usepackage{times}
\usepackage{latexsym}
\usepackage{booktabs}
\usepackage{titlesec}
\usepackage{todonotes}

\usepackage{url}

\usepackage{bm}
\usepackage{physics}
\usepackage{bbm}
\usepackage{amsmath}
\usepackage{amsfonts,amssymb}
\DeclareMathOperator*{\argmax}{argmax}
\DeclareMathOperator*{\argmin}{argmin}

\usepackage{multirow}
\usepackage{ulem}

\usepackage[T1]{fontenc}

\usepackage[utf8]{inputenc}

\usepackage{microtype}
\usepackage{mathtools}
\usepackage{nicefrac}
\usepackage{enumitem}
\usepackage{graphicx}

\usepackage{listings}

\usepackage[frozencache,cachedir=.]{minted}
\usepackage[nameinlink,capitalize,noabbrev]{cleveref}

\definecolor{Red}{rgb}{1.0,0.0,0.3}
\definecolor{Green}{rgb}{0,0.6,0}
\definecolor{Blue}{rgb}{0,0.3,1.0}
\definecolor{Grey}{rgb}{0.6,0.6,0.6}

\makeatletter
    \newcommand\crulefill{\leavevmode
    \begingroup 
    \setlength{\dimen@}{0.5ex}
    \addtolength{\dimen@}{0.4pt}
    \leaders\hrule height\dimen@ depth -0.5ex \hfill
    \endgroup
    \kern\z@}

\title{Tokenization with Factorized Subword Encoding}

\author{
    David Samuel \and Lilja Øvrelid \vspace{0.5em}\\
    University of Oslo, Language Technology Group
}

\begin{document}
\maketitle
\begin{abstract}
In recent years, language models have become increasingly larger and more complex. However, the input representations for these models continue to rely on simple and greedy subword tokenization methods. In this paper, we propose a novel tokenization method that factorizes subwords onto discrete triplets using a VQ-VAE model. The effectiveness of the proposed tokenization method, referred to as the \textsc{factorizer}, is evaluated on language modeling and morpho-syntactic tasks for 7 diverse languages. Results indicate that this method is more appropriate and robust for morphological tasks than the commonly used byte-pair encoding (BPE) tokenization algorithm.
\vspace{1em}
\end{abstract}

\section{Introduction}

\definecolor{word}{RGB}{170, 89, 171}
\definecolor{linguistics}{RGB}{210, 235, 45}
\definecolor{words}{RGB}{59, 89, 171}
\definecolor{word-}{RGB}{35, 89, 171}
\definecolor{-lessness}{RGB}{195, 161, 161}

A typical subword tokenizer consists of a vocabulary of 10\,000s of subwords, each of them mapped onto a single and independent index. Instead, we propose a method that learns to project subwords onto triplets of $256$ indices:
\begin{align*}
\footnotesize\textit{\textvisiblespace ${melon}$\textvisiblespace} &\to \footnotesize\textrm{$[30, {\color{Red}{255}}, {\color{Red}{209}}]$} 
.\\[0.5em]
\intertext{This mapping is learned by a vector-quantized variational auto-encoder \citep[VQ-VAE;][]{NIPS2017_7a98af17} from a large word-frequency list, resulting in a projection where orthographically different words use different indices and similar words share similar indices:}
\footnotesize\textit{\textvisiblespace $melons$\textvisiblespace} &\to 
\footnotesize\textrm{$\left[261, {\color{Red}{255}}, {\color{Red}{209}}\right]$},\\
\footnotesize\textit{\textvisiblespace $water$\textvisiblespace} &\to 
\footnotesize\textrm{$[96, {\color{Blue}{235}}, {\color{Blue}{109}}]$}.
\\[0.5em]
\intertext{Modeling this projection with a VQ-VAE also automatically gives the estimated probability of every subword given an index triplet. Maximizing the joint probability allows for an optimal subword tokenization of words not contained in the vocabulary.}
\footnotesize\textit{\textvisiblespace $water\,\,|\,\,melon$\textvisiblespace} &\to 
\footnotesize\textrm{
$[208, {\color{Blue}{235}}, {\color{Blue}{109}}
]$, $[45, {\color{Red}{255}}, {\color{Red}{209}}]$}.\\
\end{align*}

\noindent
In this paper, we present \textsc{factorizer}, a subword encoding method that serves as a drop-in replacement for any subword tokenizer used in modern NLP pipelines. We release the source code, trained models and ready-to-use tokenizers online.\footnote{\url{https://github.com/ltgoslo/factorizer}} Our approach demonstrates the following advantages:

\begin{enumerate}
    \item \textbf{Improved performance.} We evaluate the performance of our \textsc{factorizer} on masked language models of seven linguistically diverse languages: Arabic, Chinese, Czech, English, Norwegian, Scottish Gaelic and Turkish. These models are evaluated on part-of-speech tagging, dependency parsing, and lemmatization tasks and demonstrate a substantial improvement in performance.
    
    \item \textbf{Increased robustness.} Additionally, we investigate the robustness of our \textsc{factorizer} to random noise during inference as well as the robustness to data scarcity during pretraining. We measure performance with increasing levels of noise and data scarcity and demonstrate that our \textsc{factorizer} improves robustness to these factors.
    
    \item \textbf{More effective use of parameters.} Traditional subword tokenizers require large vocabularies to cover most word forms, which results in a substantial portion of learnable parameters being consumed by the subword embedding layer. For example, the embedding layer of BERT\textsubscript{\textit{base}} uses more than 20\% of its parameters, totaling over 23 million. In contrast, the memory footprint of our \textsc{factorizer} embedding is substantially lower as it only requires about 0.6 million parameters.\footnote{It uses embedding matrices for $3 \times 256$ indices, where each embedding vector has length 768, as in BERT\textsubscript{\textit{base}}.} The remaining parameters can be then allocated more effectively in self-attention layers.

\end{enumerate}


\section{Background: VQ-VAE}

\begin{figure}[!t]
    \includegraphics[trim={3em 0 0 0}, clip, width=\columnwidth]{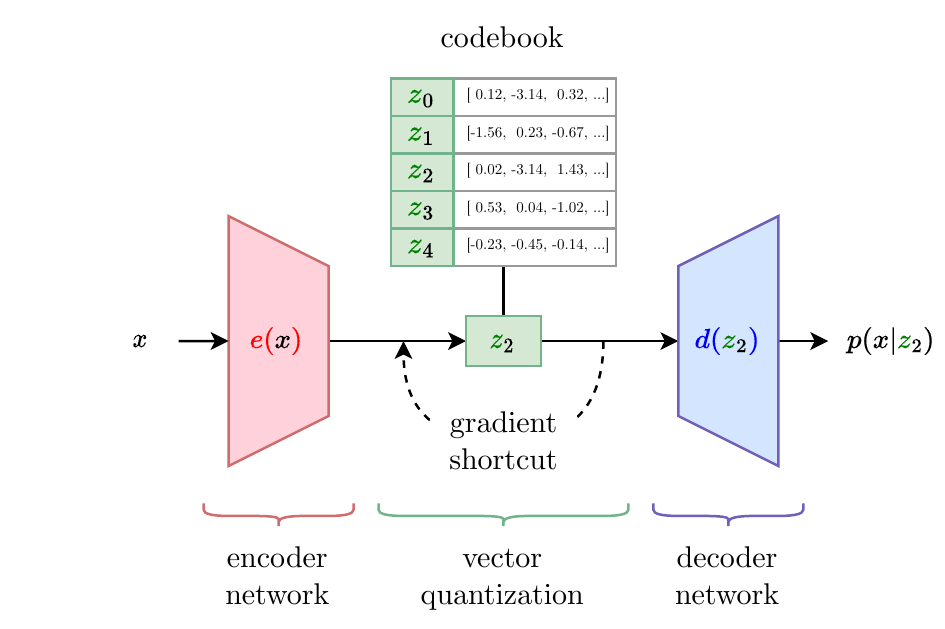}
    \caption{The main components of a VQ-VAE model. The latent vector $\color{Red}e(\color{black}\bm{x}\color{Red})$ is quantized into the 2\textsuperscript{nd} codebook vector $\color{Green}\bm{z}\color{black}_2$ and then the decoder tries to reconstruct the original input as $\color{Blue}d(\color{Green}\bm{z}\color{black}_2\color{Blue})$. We use the codebook size $K=5$ in this illustrative example, i.e. all vectors are clustered into five  possible values.}
    \label{fig:vqvae}
\end{figure}

In this paper, we propose a novel tokenizer that utilizes a vector-quantized variational auto-encoder \citep[VQ-VAE;][]{NIPS2017_7a98af17} as its central component. VQ-VAE is a powerful technique for learning \textit{discrete} latent variables that can encode words and reconstruct them back to their original form. This process can be broken down into three main steps, as illustrated in \cref{fig:vqvae}:


\begin{enumerate}\itemsep0em 
    \item \textbf{The encoder} maps the data samples (subwords) $\bm{x}$ to continuous latent vectors $\color{Red}e(\color{black}\bm{x}\color{Red})$.
    \item \textbf{The codebook} is made up of $K$ codebook vectors $\color{Green}\bm{z}_{\color{black}k}\color{black}, k \in 1 \dots K$. Each latent vector $\color{Red}e(\color{black}\bm{x}\color{Red})$ is \textit{quantized} to its nearest codebook vector $\color{Green}\bm{z}_{\color{black}k}$, effectively mapping each input sample $\bm{x}$ onto a discrete variable $k$. The encoder and codebook downsample and compress the information in $\bm{x}$, serving as an \textit{information bottleneck} \cite{tishby99information}.
    \item \textbf{The decoder} reconstructs the symbols $\color{Green}\bm{z}$ back into the original input distribution, modeling the distribution $p(\bm{x}|\color{Green}\bm{z}\color{black})$ as $\color{Blue}d(\color{Green}\bm{z}\color{Blue})$.
\end{enumerate}

\paragraph{Gradient backpropagation.} The model is optimized jointly with the backpropagation algorithm. However, special attention must be given to the codebook quantization $\color{Green}q$ as this operation is not differentiable:
\begin{equation}\label{eq:quantization}
\begin{split}
\color{Green}q\!\left(\color{Red}e(\color{black}\bm{x}\color{Red})\color{Green}\right)\color{black} &= \color{Green}\bm{z}_{\color{black}k}\color{black},\,\,  \textrm{where} \\
k &= \argmin_i{\lVert \color{Red}e(\color{black}\bm{x}\color{Red})\color{black} - \color{Green}\bm{z}\color{black}_i \rVert_2} \\
\end{split}
\end{equation}
During the forward phase, the quantized codebook vectors are simply passed to the decoder. However, during the backward phase, the gradient of the loss function $\nabla_{\!\color{Blue}d\color{black}} \mathcal{L}$ is passed directly from the decoder to the encoder. This technique, known as \textit{straight-through gradient estimation}, is possible because the output of the encoder and the input of the decoder share the same latent space. The output $\color{Red}e(\color{black}\bm{x}\color{Red})$ is sufficiently similar to the decoder input $\bm{z}$, such that the gradient carries information about how to change $\color{Red}e(\color{black}\bm{x}\color{Red})$ to lower the reconstruction loss.

\paragraph{Loss terms.} In addition to the standard \textit{reconstruction loss} $\mathcal{L}_r = \textrm{log}\,p(\bm{x}|\color{Green}\bm{z}\color{black})$, which measures the auto-encoding performance, the VQ-VAE model incorporates two auxiliary loss terms that align the encoder outputs $\color{Red}e(\color{black}\bm{x}\color{Red})$ with the nearest codebook vectors. Specifically, the \textit{codebook loss} is defined as $\mathcal{L}_{\color{Green}q} = \lVert \textrm{sg}(\color{Red}e(\color{black}\bm{x}\color{Red})\color{black}) - \color{Green}\bm{z}\color{black}_k \rVert_2$ and the \textit{commitment loss} as $\mathcal{L}_{\color{Red}e} = \lVert \color{Red}e(\color{black}\bm{x}\color{Red})\color{black} - \textrm{sg}(\color{Green}\bm{z}\color{black}_k) \rVert_2$, where `sg' is the stop gradient operation and $\color{Green}\bm{z}\color{black}_k$ is the codebook vector defined in \cref{eq:quantization}. The overall loss of a VQ-VAE model is computed as:
\[\mathcal{L} = \mathcal{L}_r + \mathcal{L}_{\color{Green}q} + \beta \mathcal{L}_{\color{Red}e},\]
where $\beta$ is a hyperparameter that balances the focus between reconstruction and codebook alignment.

\paragraph{Codebook EMA.} An alternative approach to updating the codebook, proposed by \newcite{NIPS2017_7a98af17}, is to update it as an exponential moving average (EMA) of the latent variables $\color{Red}e(\color{black}\bm{x}\color{Red})\color{black}$ -- instead of incorporating the codebook loss $\mathcal{L}_{\color{Green}q}$. This update consists of updating two variables: the codebook usage counts $c_k$ and the codebook vectors $\color{Green}\bm{z}_{\color{black}k}$, where $\lambda$ is the EMA decay hyperparameter:
\begin{equation*}
\begin{split}
c_k & \leftarrow \lambda c_k + (1 - \lambda)\sum_i{\mathbbm{1}_{\left[\color{Green}q(\color{Red}e(\color{black}\bm{x}_i\color{Red})\color{Green})\color{black}\,=\,\color{Green}\bm{z}\color{black}_k\right]}} \\
\color{Green}\bm{z}\color{black}_k & \leftarrow \lambda \color{Green}\bm{z}\color{black}_k +\frac{(1 - \lambda)}{c_k}\sum_i{\color{Red}e(\color{black}\bm{x}_i\color{Red})\color{black}\mathbbm{1}_{\left[\color{Green}q(\color{Red}e(\color{black}\bm{x}_i\color{Red})\color{Green})\color{black}\,=\,\color{Green}\bm{z}\color{black}_k\right]}}
\end{split}
\end{equation*}
This approach leads to more stable training \citep{pmlr-v80-kaiser18a} and it allows us to mitigate the \textit{codebook collapse}, which occurs when the usage count of a vector drops to zero and the vector is then never updated \citep{pmlr-v80-kaiser18a}. Thus, whenever a codebook vector $\color{Green}\bm{z}\color{black}_i$ has usage $\bm{c}_i$ lower than $c_{\textrm{min}}$, it is reassigned to a random latent vector $\color{Green}\bm{z}\color{black}_i \leftarrow \color{Red}e(\color{black}\bm{x}_j\color{Red})$ and the usage count is reset $c_i \leftarrow 1$, similar to \newcite{NEURIPS2020_309fee4e} or \newcite{dhariwal2020jukebox}.

\newpage

\section{\textsc{Factorizer}}
\label{sec:colorizer}

We utilize the VQ-VAE architecture to train a model capable of mapping discrete triplet representations $\color{Green}\bm{z}$ to subword strings $\bm{w}$ (and vice versa). Furthermore, we employ the VQ-VAE decoder to estimate the probabilities of the subword strings $p(\bm{w}|\color{Green}\bm{z}\color{black})$. After the model is trained, we infer its vocabulary, consisting of a set of tuples $\langle \bm{w}_i, \color{Green}\bm{z}\color{black}_i, \textrm{log}\,p(\bm{w}_i|\color{Green}\bm{z}\color{black}_i) \rangle$. Finally, we use this vocabulary to perform optimal (in regards to the log-probabilities) subword tokenization.

\subsection{VQ-VAE subword encoding}

\paragraph{Training data.} The auto-encoder is trained on a word-frequency list obtained by word-tokenizing a large text corpus. Let us denote the frequency of word $\bm{w}$ by $f_{\bm{w}}$. Note that while this data representation simplifies the model by discarding any contextual information, it still enables proper estimation of $p(\bm{w}_i|\color{Green}\bm{z}\color{black}_i)$ by following the word frequencies $f_{\bm{w}}$. 

\paragraph{Word representation.} In this study, words are represented as sequences of UTF\nobreakdash-8 bytes. To ensure proper handling of word boundaries, the word sequences start with a special ``beginning-of-word'' token and end with an ``end-of-word'' token; both tokens are depicted as the symbol `\textvisiblespace' in this text.

\paragraph{Data sampling.} In theory, the correct way of sampling the training data is to directly follow the frequencies $f_{\bm{w}}$. However, in practice, the distribution of words in a natural language follows Zipf's law, resulting in a skewed distribution that collapses the training process. To address this issue, we sample the data according to a more balanced distribution
\[
p_{\textrm{sample}}(w) \propto \frac{f_{\bm{w}}}{\textrm{log}(f_{\bm{w}} + 1)}.
\]
To compensate for this alteration and accurately model the true word distribution, we incorporate the denominator into the loss function by weighting it as follows:
\[
\mathcal{L} = \sum_{\bm{w}}{\textrm{log}(f_{\bm{w}} + 1)\,\mathcal{L}(\bm{w})}.
\]

\paragraph{Subword splitting.} Up to this point, we have only discussed word encoding. To also encode \textit{sub}words, we randomly split some of the sampled words into subwords. Specifically, as the more frequent words should be split less frequently, we keep the word as-is with the probability
\[
p_{\textrm{not-split}}(w) = \frac{\textrm{log}(f_{\bm{w}} + 1)}{\textrm{max}_{\bm{x}}\,\textrm{log}(f_{\bm{x}} + 1)}.
\]

\paragraph{Factorized codebooks.} To capture fine-grained information about the characters inside words, we employ separate codebooks, each within a unique latent space. In addition, to further reduce the issue of codebook collapse, each codebook consists only of $K=256$ codebook vectors.\footnote{Note that this is reminiscent of the standard 24-bit RGB color encoding, where every codebook triplet can be viewed as an RGB color. \textsc{Factorizer} essentially projects subwords into a color space, see \cref{sec:color} for more details.}

\paragraph{Backbone architecture.} The auto-encoder model is based on an encoder-decoder transformer architecture \citep{NIPS2017_3f5ee243} with three quantization bottlenecks. Specifically, we first pad the byte-tokens $\bm{w}$ with special \texttt{R}, \texttt{G}, \texttt{B} prior tokens, then embed and encode each sequence with a transformer encoder. The contextualized embedding vectors for the first three tokens serve as the encoding $\color{Red}e(\color{black}\bm{w}\color{Red})$. These three vectors are then quantized and prepended to the subword byte-tokens $\bm{w}$, which are finally input into the autoregressive transformer decoder $\color{Blue}d(\color{Green}\bm{z}\color{black}, \bm{w}\color{Blue})$, modeling $p(\bm{w}|\color{Green}\bm{z}\color{black})$.

\paragraph{Training details.} All VQ-VAE models in this study utilize a transformer with 6 layers and 256 hidden size for both the encoder and decoder. The models are trained for 50\,000 steps with batch size of 4\,096 and optimized with adaptive sharpness-aware minimization \citep[ASAM;][]{pmlr-v139-kwon21b} to ensure better generalization to unseen words, with AdamW \citep{loshchilov2018decoupled} as the underlying optimizer. To further improve generalization, we calculate the exponential moving average of the parameters (with a decay of 0.999) and use this average for inference. For more details about the transformer architecture and hyperparameters, please refer to \newcite{samuel-etal-2023-trained} and \cref{sec:hyperparameters}.

\paragraph{Vocabulary construction.} The VQ-VAE is not used directly for tokenization as that would greatly slow it down. Instead, we infer its static vocabulary by iterating through all $256^3$ instances of the RGB codebooks. The subword vocabulary is decoded as
\[
\mathcal{W} = \left\{\argmax_{\bm{w}} p(\bm{w}|\color{Green}\bm{z}\color{black})\,\,\middle|\,\,\color{Green}\bm{z}\color{black} \in [256]^3\!,\,\,p(\color{Green}\bm{z}\color{black})\!>\!0\right\}\!,
\]
where the prior distribution $p(\color{Green}\bm{z}\color{black})$ is calculated by counting the usage of all codebook triplets throughout training. Finally, the full vocabulary $\mathcal{V}$ is a set of tuples $\langle \bm{w}_i, \color{Green}\bm{z}\color{black}_i, \textrm{log}\,p(\bm{w}_i|\color{Green}\bm{z}\color{black}_i) \rangle$, where $\bm{w}_i \in \mathcal{W}$ and $\color{Green}\bm{z}\color{black}_i = \argmax_{\color{Green}\bm{z}} p(\bm{w}_i|\color{Green}\bm{z}\color{black})$.

\begin{figure*}[!ht]
    \includegraphics[width=\textwidth]{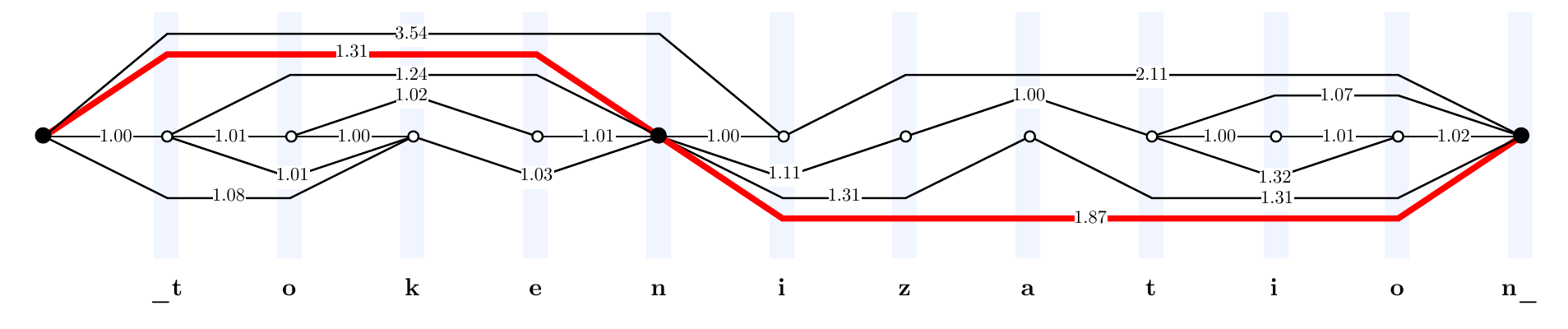}
    \caption{Diagram showing an illustrative subset of the state space when searching for the optimal split of word \textit{``tokenization''}. Every subword $\bm{w}$ is associated with a weighted directed edge in the state graph of value score($\bm{w}$); the optimal tokenization is then equivalent to finding the shortest path between the leftmost and the rightmost node.}
    \label{fig:tokenization}
\end{figure*}

\subsection{Subword tokenizer}

\begin{figure}[!t]
    \includegraphics[width=\columnwidth]{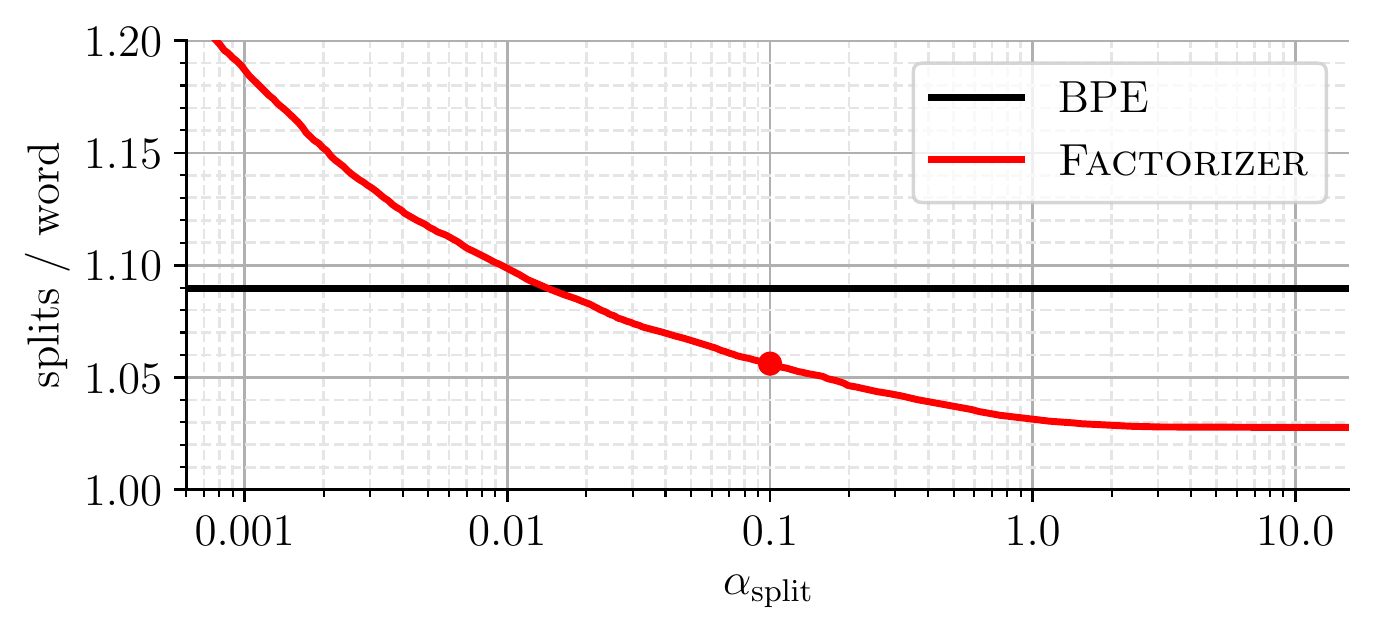}
    \caption{The influence of the splitting parameter $\alpha_{\textrm{split}}$ on the amount of splits per word, as calculated on the \texttt{en-ewt} dataset from Universal Dependencies. The BPE tokenizer has vocabulary size of 32\,768.}
    \label{fig:alpha-splits}
\end{figure}

\paragraph{Optimal split search.} After inferring the vocabulary $\mathcal{V}$, we can search for the optimal tokenization of a word $\bm{x}$ into subwords $\bm{w}_1,\bm{w}_2,\dots \bm{w}_k$:
$$
\textrm{tokenize}(\bm{x}) =\hspace{-0.6em}\argmin_{\bm{w}_1\!\bm{w}_2\dots \bm{w}_k\,=\,\bm{x}}\,\,{\sum_{i=1}^k{\textrm{score}(\bm{w}_i)}},\hspace{0.5em}
$$
where for each subword $\bm{w}_i$ from the vocabulary triplets $\langle \bm{w}_i, \color{Green}\bm{z}\color{black}_i, \textrm{log}\,p(\bm{w}_i|\color{Green}\bm{z}\color{black}_i) \rangle \in \mathcal{V}$, its score is
$$
\textrm{score}(\bm{w}_i) = -\textrm{log}\,p(\bm{w}_i|\color{Green}\bm{z}_i\color{black}) + \alpha_{\textrm{split}}.\hspace{1.4em}
$$The parameter $\alpha_{\textrm{split}}$ allows for a smooth change of the amount of splits per word, as shown in \cref{fig:alpha-splits}. We use $\alpha_{\textrm{split}}=0.1$ for all experiments in this work. 

The tokenize function is implemented via the shortest path search in a tokenization graph, as illustrated in \cref{fig:tokenization}. Specifically, the search uses Dijkstra's algorithm \citep{dijkstra1959note} and the forward edges from each node are efficiently iterated with the prefix search in a directed acyclic word graph \citep[DAWG;][]{BLUMER198531} of all subwords from $\mathcal{V}$. A simplified pseudocode of the search algorithm is given below, the optimal tokenization is then obtained by reverse iteration of the returned \texttt{prev} dictionary.

\begin{minted}[escapeinside=||, xleftmargin=0em, linenos=false, fontsize=\scriptsize]{python}
def search(|\bfseries{\color{Grey}word}|, |\bfseries{\color{Grey}dawg\_trie}|):
    |\bfseries{\color{Grey}Q}|.append(|\bfseries{\color{Grey}word}|)        |\textit{\color{Grey}priority queue of suffixes}|
    |\bfseries{\color{Grey}cost}|[|\bfseries{\color{Grey}word}|] = 0.0    |\textit{\color{Grey}cost returns +inf by default}|
    |\bfseries{\color{Grey}prev}|[|\bfseries{\color{Grey}word}|] = none |       \textit{\color{Grey}pointers to predecesors}|

    while |\bfseries{\color{Grey}Q}| |\texttt{is not empty}|:
        |\bfseries{\color{Grey}suffix}| = suffix |\texttt{from}| |\bfseries{\color{Grey}Q}| |\texttt{with}| minimal |\bfseries{\color{Grey}cost}|
        remove |\bfseries{\color{Grey}suffix}| |\texttt{from}| |\bfseries{\color{Grey}Q}|
        if |\bfseries{\color{Grey}suffix}| |\texttt{is}| empty: break

        for every |\bfseries{\color{Grey}prefix}| of |\bfseries{\color{Grey}suffix}| |\texttt{in}| |\bfseries{\color{Grey}dawg\_trie}|:
            |\bfseries{\color{Grey}new\_cost}| = |\bfseries{\color{Grey}cost}|[|\bfseries{\color{Grey}suffix}|] + score(|\bfseries{\color{Grey}prefix}|)
            |\bfseries{\color{Grey}new\_suffix}| = |\bfseries{\color{Grey}suffix}|[len(|\bfseries{\color{Grey}prefix}|):]
            if |\bfseries{\color{Grey}new\_cost}| < |\bfseries{\color{Grey}cost}|[|\bfseries{\color{Grey}new\_suffix}|]:
                |\bfseries{\color{Grey}Q}|.append(|\bfseries{\color{Grey}new\_suffix}|)
                |\bfseries{\color{Grey}cost}|[|\bfseries{\color{Grey}new\_suffix}|] = |\bfseries{\color{Grey}new\_cost}|
                |\bfseries{\color{Grey}prev}|[|\bfseries{\color{Grey}new\_suffix}|] = |\bfseries{\color{Grey}min\_suffix}|

    return |\bfseries{\color{Grey}prev}|
\end{minted}

\paragraph{Sampling.} Our formulation of the tokenizer allows for a straightforward modification for sampling different subword splits. This is similar to BPE dropout \citep{provilkov-etal-2020-bpe}, a powerful regularization technique. 
The sampling works by modifying the score function as follows, making sure that all scores are non-negative so that the correctness of the search method holds:
\begin{equation*}
\begin{split}
\textrm{score}(\bm{w}_i) &= -\textrm{log}\,p(\bm{w}_i|\color{Green}\bm{z}_i\color{black}) + \alpha_{\textrm{split}} + |\bm{w}_i|\textrm{exp}(\epsilon) \\
\epsilon &\sim \mathcal{N}(0, \sigma_{\textrm{sample}}^2)
\end{split}
\end{equation*}
We set the parameter $\sigma_{\textrm{sample}}$ to 0.02 in all sampling experiments.

\subsection{Application}

\paragraph{Embedding layer.} The three \textsc{factorizer} subword indices are embedded with separate embedding layers, summed and transformed with a GeLU non-linearity. In the end, we get a single embedding vector for each subword, which makes it a simple drop-in replacement for standard embedding layers.

\paragraph{Averaging.} The random subword sampling can be utilized for more robust inference by sampling multiple tokenizations of each data instance and averaging the predictions for each of them.


\clearpage
\section{Experiments}

The effectiveness of \textsc{factorizer} is experimentally verified on a typologically diverse set of languages. We train masked language models on each language and then evaluate it on morpho-syntactic tasks from the Universal Dependencies (UD) treebanks \citep{nivre-etal-2016-universal}. Additionally, we demonstrate that the \textsc{factorizer}-based language models exhibit greater robustness to noise and superior performance in low-resource settings. We present four sets of experiments in this section:
\begin{enumerate}\itemsep0em
    \item In order to provide some initial observations, we evaluate simple parsing models that do not rely on any pretrained embeddings.
    \item Then, we ablate different tokenizer configurations with English language modeling and UD finetuning.
    \item As the main experiment, we pretrain and finetune language models on 7 typologically diverse languages.
    \item Lastly, controlled experiments on English examine the robustness of our method to noise and data scarcity.
\end{enumerate}

\paragraph{Languages.}

In total, our method is evaluated on 7 different languages to investigate its performance on different morphological typologies. For the most part, we follow \newcite{vania-lopez-2017-characters} in their selection of languages according to the traditional categories of morphological systems. The text corpus for each language is drawn from the respective part of the mC4 corpora \citep{xue-etal-2021-mt5}, unless specified otherwise. Note that we use the same corpus for training \textsc{factorizer} models (where we extract a word-frequency list for every language) as for training language models. The languages chosen for evaluation are:

\begin{itemize}\itemsep0em
    \item \textbf{Arabic} -- The first distinct language type used for evaluation are \textit{introflexive} languages. We decided to use Arabic as a representative of introflexive languages because it is arguably the most wide-spread language in this class. Arabic also tests how a tokenizer can handle a non-Latin script.
    \item \textbf{Chinese} -- A purely \textit{analytical} language; evaluation on Chinese tests performance on a completely different morphological typology than the other 6 languages.
    \item \textbf{Czech} -- An Indo-European language which exhibits fusional features and a \textit{rich system of morphology}, typical for all Slavic languages.
    \item \textbf{English} -- In principle, also a \textit{fusional} language, however, one of the most analytical languages in this class. 
    The English corpus is made of the publicly available replications of BookCorpus and OpenWebText (because of the unnecessarily large size of the English part of mC4 for our purposes).\footnote{\url{https://the-eye.eu/public/AI/pile\_preliminary\_components/books1.tar.gz}}\footnote{\url{https://openwebtext2.readthedocs.io}}
    \item \textbf{Norwegian} -- Another Germanic language, typologically close to English, but has different morphological properties, in particular highly productive use of \textit{morphological compounding}, which makes it a fitting choice for evaluation of tokenizers.
    \item \textbf{Scottish Gaelic} -- Also a fusional language. In order to evaluate performance on \textit{low-resource} languages, this language, which has about 57\,000 fluent speakers, was chosen as it is the smallest language from mC4 with a full train-dev-test split in Universal Dependencies.
    \item \textbf{Turkish} -- As an \textit{agglutinative} language, Turkish is on the other end of the `morphemes-per-word spectrum' than an analytical language like Chinese.
\end{itemize}

\paragraph{BPE tokenizer baseline.}

We compare the performance of \textsc{factorizer} with the byte-pair encoding compression algorithm \citep[BPE;][]{10.5555/177910.177914, sennrich-etal-2016-neural} as the most commonly used subword tokenizer. Following recent language modeling improvements \citep{radford2019language}, we use BPE directly on UTF-8 bytes instead of unicode characters -- similarly to \textsc{factorizer}, which also utilizes UTF-8 bytes as the atomic unit. The training itself uses the open implementation from \texttt{tokenizers} library\footnote{\url{https://github.com/huggingface/tokenizers}} and utilizes the same text corpora as the corresponding \textsc{factorizer} models. The choice of BPE also allows for a comparison between \textsc{factorizer} sampling and BPE dropout \citep{provilkov-etal-2020-bpe}.

\subsection{Experiment 1: UD parser `from scratch'}
\label{sec:from-scratch}

Before training large language models and evaluating the tokenizers in a more resource-intensive setting, we motivate our proposed method on a simple model that does not utilize any pretrained word embeddings.

\paragraph{UD parser architecture.} We base this first experiment on UDPipe 2, a publicly available model for PoS tagging, lemmatization and dependency parsing by \newcite{straka-etal-2019-udpipe}. We generally follow the original architecture, only replacing its word embeddings -- UDPipe 2 utilizes a combination of pretrained contextualized embeddings, static embeddings and character embeddings -- instead, we use only randomly initialized subword embeddings and learn them from scratch; the final contextualized representations for the subwords are average-pooled for each word-token to obtain word-level representations. Otherwise we use the same joint modeling approach based on bidirectional LSTM layers \citep{HochSchm97} followed by individual classification heads for \texttt{UPOS}, \texttt{XPOS} and \texttt{UFeats} tagging, a classification head for relative lemma rules and a final head for dependency parsing \citep{dozat2017deep}.

\begin{figure}[!t]
    \includegraphics[trim={0 0 0 0em},width=0.945\columnwidth]{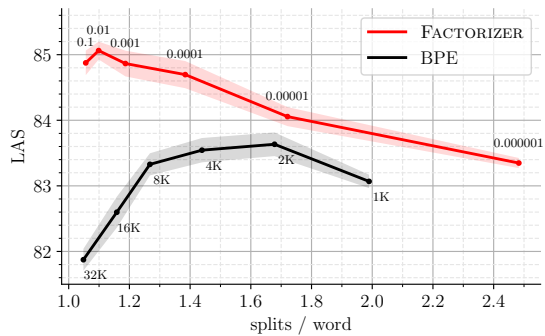}
    \caption{The LAS scores on the development split of \texttt{en-ewt}. We evaluate multiple `from scratch' models trained on top of tokenizers with different amounts of subword splits-per-word. \textsc{Factorizer} can change this value by altering the $\alpha_{\textrm{split}}$ parameter, while BPE influences this value indirectly by altering its vocabulary size. The lightly-colored regions illustrate the standard deviation of 5 runs with different random seeds.
    }
    \label{fig:nor-splits}
\end{figure}

\begin{figure}[!t]
    \includegraphics[width=\columnwidth]{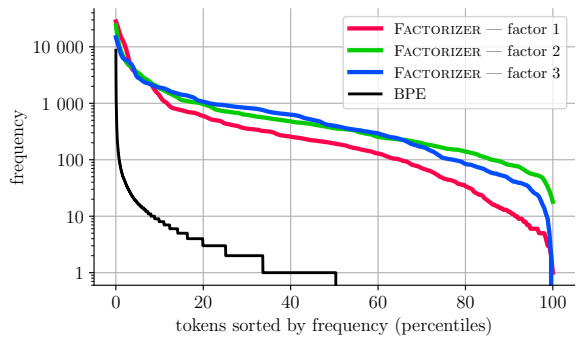}
    \caption{The number of occurrences of each subword index in \texttt{en-ewt}. The BPE subwords clearly follow the Zipf's law, with 90\% of them appearing less than 10 times. On the other hand, the factorized indices are more evenly distributed, which makes it easier to learn their embeddings with less data.}
    \label{fig:histogram}
\end{figure}

\paragraph{Results.} The amount of splits per word has a noticeable impact on the performance, we therefore include this dimension in the evaluation, visualized in \cref{fig:nor-splits}. This shows that \textsc{factorizer} is clearly a Pareto-optimal tokenizer in this comparison. We conjecture that this is caused by its ability to learn a usable embedding for less frequent, or even out-of-vocabulary subwords, as illustrated in \cref{fig:histogram}.

\subsection{Experiment 2: Ablation study with English language modeling}
\label{sec:ablation}

In order to evaluate the performance of different BPE and \textsc{factorizer} configurations, we conduct a comparative study on English. 

\paragraph{Language model pretraining.} In practice, the capabilities of a UD parser can be massively improved by utilizing contextualized embeddings from large language models that have been trained on vast amounts of data. In light of this, the main focus of this experimental section is to evaluate the impact of the \textsc{factorizer} on language models and their downstream performance.

We follow the training budget and parameters of the original BERT\textsubscript{base} \citep{devlin-etal-2019-bert} for pretraining the language models, the BPE-based model also uses BERT's vocabulary size of 32K. But unlike BERT\textsubscript{base}, we establish our models on the more efficient LTG-BERT architecture \citep{samuel-etal-2023-trained}, which enhances the models with some recent improvements, such as NormFormer layer normalization \citep{shleifer2022normformer}, disentangled attention with relative positional encoding \citep{he2021deberta} or span masking \citep{joshi-etal-2020-spanbert}. In order to reduce training time, pretraining is paralellized over 128 GPUs, the total batch size is increased to 8\,192 and the amount of steps is reduced to 31\,250, matching the training budget of BERT\textsubscript{base}. Please refer \citet{samuel-etal-2023-trained} for details on the architecture and \cref{sec:hyperparameters}  for the set of all hyperparameters.

In order to make the comparison between both tokenizers fair, we approximately match the number of trainable parameters in all language models. \textsc{Factorizer} requires relatively small embedding layers, so we move its `parameter budget' to 3 additional transformer layers -- then the BPE-based models use 110.8M parameters while the \textsc{factorizer}-based models use 108.1M parameters in total. The effect of this choice is evaluated in \cref{sec:ablation-full}.

\paragraph{UD finetuning.} We employ the same model as in \cref{sec:from-scratch}, only replacing the LSTM layers with a convex combination of hidden layers from a pretrained language model, similarly to UDify \citep{kondratyuk-straka-2019-75}. Then we finetune the full model on a UD treebank. The detailed hyperparameter choice is given in the \cref{sec:hyperparameters}.

\paragraph{Results.} The comparison is presented in \cref{tab:comparative}. On average, \textsc{factorizer} outperforms BPE, especially in lemmatization. The results show that both sampling and averaging techniques (discussed in \cref{sec:colorizer}) improve performance, with the \textsc{factorizer} particularly benefiting from these features.

\subsection{Experiment 3: Multi-lingual evaluation.} Finally, we train 7 BPE-based and \textsc{factorizer}-based language models on the selected languages and finetune them on the respective UD treebanks.

\paragraph{Results.} The results, displayed in \cref{tab:multilingual}, indicate that \textsc{factorizer} clearly achieves better performance than BPE on average. Furthermore, the \textsc{factorizer} is consistently more effective on lemmatization (with statistical significance), suggesting that the factorized subwords are indeed able to carry information about the characters inside the subword-units. 

Interestingly, \textsc{factorizer} does not always perform better on Czech and Turkish, even though we hypothesized that these languages should benefit from a better character information given their rich morphology. Neither BPE or \textsc{factorizer} is an overall better choice for these languages.

While the results for dependency parsing in isolation (as measured by \texttt{UAS} and \texttt{LAS}) are more mixed, the parsing results which further take into account morphology (\texttt{MLAS}) and lemmatization (\texttt{BLEX}) largely favor the factorized subword encoding. \textsc{Factorizer}  outperforms BPE (with statistical significance) in 6 out of 7 languages on \texttt{BLEX} scores.

{\renewcommand{\arraystretch}{1.33}
\begin{table}[t]
\resizebox{\columnwidth}{!}{%
\begin{tabular}{@{}llll@{}}
\toprule
\textbf{Model}                                                                & \textbf{AllTags}  & \textbf{Lemmas}   & \textbf{LAS}      \\ \midrule
BPE                                     & 
96.05$^{\pm0.06}$ & 98.03$^{\pm0.03}$ & 92.06$^{\pm0.08}$ \\
BPE + sampling                                  & 
96.21$^{\pm0.04}$ & 98.25$^{\pm0.01}$ & 92.22$^{\pm0.05}$
\\
BPE + sampling + averaging                                  & 
96.24$^{\pm0.04}$ & 98.29$^{\pm0.04}$ & 92.22$^{\pm0.12}$ \\[0.75em]
\textsc{Factorizer}  & 
96.06$^{\pm0.05}$ & 98.01$^{\pm0.01}$ & 92.25$^{\pm0.13}$ \\
\textsc{Factorizer} + sampling  &
96.29$^{\pm0.08}$ & 98.40$^{\pm0.02}$ & 92.29$^{\pm0.07}$ \\
\textsc{Factorizer} + sampling + averaging & 
\textbf{96.34}$^{\pm0.05}$ & \textbf{98.48}$^{\pm0.03}$ & \textbf{92.39}$^{\pm0.14}$ \\ \bottomrule
\end{tabular}%
}
\caption{Comparison of BPE and \textsc{factorizer} configurations. The metrics are measured on the development split of \texttt{en-ewt}. We show the mean and standard deviation statistics over 5 random seeds, the best results are displayed in bold. Full results are shown in \cref{tab:full-comparison}.}
\label{tab:comparative}
\end{table}
}

{\renewcommand{\arraystretch}{1.34}
\begin{table*}[!t]
\resizebox{\textwidth}{!}{%
\begin{tabular}{@{}l@{\hspace{4em}}lll@{\hspace{3em}}l@{\hspace{3em}}ll@{\hspace{3em}}ll@{}}
\toprule
\textbf{Model}  & \textbf{UPOS}     & \textbf{XPOS}      & \textbf{UFeats}   & \textbf{Lemmas}   & \textbf{UAS}      & \textbf{LAS}      & \textbf{MLAS}     & \textbf{BLEX}     \\ \midrule

\multicolumn{5}{@{}l}{\raisebox{0.5ex}{\footnotesize Arabic (\texttt{ar-padt})}} \\
\hspace{1em}UDPipe 2  & 97.02 & 94.38 & 94.53 & 95.31 & 88.11 & 83.49 & 74.57 & 76.13\\
\hspace{1em}Stanza & 62.06 & 54.60 & 68.34 & 49.20 & 42.72 & 39.75 & 39.07 & 38.75 \\
\hspace{1em}BPE                    & 
\textbf{97.48}$^{\pm0.02}$ & 95.84$^{\pm0.03}$ & 95.97$^{\pm0.03}$ & 95.14$^{\pm0.04}$ & 90.84$^{\pm0.07}$ & 86.50$^{\pm0.07}$ & 79.03$^{\pm0.15}$ & 79.31$^{\pm0.11}$\\ 
\hspace{1em}\textsc{Factorizer}            & 
\textbf{97.48}$^{\pm0.03}$ & \uuline{\textbf{95.90}$^{\pm0.03}$} & \uuline{\textbf{96.03}$^{\pm0.05}$} & \uuline{\textbf{95.91}$^{\pm0.04}$} & \uuline{\textbf{91.24}$^{\pm0.08}$} & \uuline{\textbf{87.05}$^{\pm0.09}$} & \uuline{\textbf{79.70}$^{\pm0.11}$} & \uuline{\textbf{80.80}$^{\pm0.11}$} \\ \midrule

\multicolumn{5}{@{}l}{\raisebox{0.5ex}{\footnotesize Chinese (\texttt{zh-gsd})}} \\
\hspace{1em}UDPipe 2  & 96.21 & 96.08 & 99.40 &  \textbf{99.99} & 87.15 & 83.96 & 78.41 & 82.59 \\
\hspace{1em}Stanza & 95.35 & 95.10 & 99.13 & \textbf{99.99} & 81.06 & 77.38 & 72.04 & 76.37 \\
\hspace{1em}BPE                    & 
\uuline{\textbf{97.14}$^{\pm0.10}$} & \uuline{\textbf{96.95}$^{\pm0.10}$} & 99.60$^{\pm0.01}$ & \textbf{99.99}$^{\pm0.00}$ & 88.32$^{\pm0.16}$ & 85.42$^{\pm0.20}$ & 80.23$^{\pm0.23}$ & 84.33$^{\pm0.17}$ \\
\hspace{1em}\textsc{Factorizer}            & 
97.00$^{\pm0.07}$ & 96.79$^{\pm0.08}$ & \uuline{\textbf{99.66}$^{\pm0.03}$} & \textbf{99.99}$^{\pm0.00}$ & \uuline{\textbf{89.34}$^{\pm0.13}$} & \uuline{\textbf{86.39}$^{\pm0.21}$} & \uuline{\textbf{81.21}$^{\pm0.16}$} & \uuline{\textbf{85.17}$^{\pm0.17}$} \\\midrule

\multicolumn{5}{@{}l}{\raisebox{0.5ex}{\footnotesize Czech (\texttt{cs-pdt})}} \\
\hspace{1em}UDPipe 2 & \textbf{99.45} & 98.47 & 98.40 & 99.28 & \textbf{95.63} & \textbf{94.23} & 90.88 & \textbf{92.50}\\
\hspace{1em}Stanza & 98.88 & 95.82 & 92.64 & 98.63 & 93.03 & 91.09 & 80.19 & 87.94 \\
\hspace{1em}BPE                    & 
\uuline{99.44$^{\pm0.01}$} & \textbf{98.54}$^{\pm0.02}$ & \textbf{98.50}$^{\pm0.02}$  & 99.26$^{\pm0.01}$ & \uuline{\textbf{95.63}$^{\pm0.04}$} & \uuline{94.15$^{\pm0.05}$} & \textbf{90.96}$^{\pm0.05}$ & 92.42$^{\pm0.04}$ \\
\hspace{1em}\textsc{Factorizer}            & 
99.39$^{\pm0.01}$ & 98.53$^{\pm0.03}$ & \textbf{98.51}$^{\pm0.03}$ & \uuline{\textbf{99.30}$^{\pm0.02}$} & 95.54$^{\pm0.04}$ & 94.10$^{\pm0.05}$ & 90.95$^{\pm0.05}$ & 92.42$^{\pm0.06}$
\\\midrule

\multicolumn{9}{@{}l}{\raisebox{0.5ex}{\footnotesize English (\texttt{en-ewt})}} \\        \hspace{1em}UDPipe 2  & 97.35 & 97.06 & 97.52 & 98.07 & 92.62 & 90.56 & 84.02 & 85.98 \\
\hspace{1em}Stanza & 96.87 & 96.67 & 97.08 & 97.82 & 90.39 & 87.94 & 80.44 & 82.57 \\
\hspace{1em}BPE                    & 
97.82$^{\pm0.04}$ & 97.65$^{\pm0.03}$ & 97.90$^{\pm0.04}$ & 97.99$^{\pm0.05}$ & 93.46$^{\pm0.09}$ & 91.63$^{\pm0.10}$ & 85.95$^{\pm0.14}$ & 87.21$^{\pm0.08}$ \\ \hspace{1em}\textsc{Factorizer}            & \uuline{\textbf{97.90}$^{\pm0.07}$} & \uuline{\textbf{97.72}$^{\pm0.06}$} & \uuline{\textbf{98.03}$^{\pm0.02}$} & \uuline{\textbf{98.29}$^{\pm0.03}$} & \uuline{\textbf{93.63}$^{\pm0.12}$} & \uuline{\textbf{91.80}$^{\pm0.15}$} & \uuline{\textbf{86.27}$^{\pm0.22}$} & \uuline{\textbf{87.90}$^{\pm0.17}$} \\\midrule
\multicolumn{5}{@{}l}{\raisebox{0.5ex}{\footnotesize Norwegian (\texttt{no-bokmaal})}} \\
\hspace{1em}UDPipe 2  & 98.61 & — & 97.68 & 98.82 & 94.40 & 92.91 & 87.59 & 89.43 \\
\hspace{1em}Stanza & 98.43 & --- & 97.71 & 98.39 & 93.39 & 91.64 & 85.60 & 87.10 \\
\hspace{1em}BPE                    & 
\uuline{\textbf{98.94}$^{\pm0.02}$} & --- & 98.21$^{\pm0.06}$ & 98.82$^{\pm0.02}$ & \textbf{95.42}$^{\pm0.07}$ & \textbf{94.13}$^{\pm0.09}$ & 89.68$^{\pm0.17}$ & 90.86$^{\pm0.10}$ \\
\hspace{1em}\textsc{Factorizer}            & 
98.90$^{\pm0.04}$ & --- & \uuline{\textbf{98.37}$^{\pm0.03}$} & \uuline{\textbf{99.19}$^{\pm0.01}$} & 95.35$^{\pm0.09}$ & 94.05$^{\pm0.08}$ & \textbf{89.85}$^{\pm0.11}$ & \uuline{\textbf{91.43}$^{\pm0.08}$} \\\midrule

\multicolumn{5}{@{}l}{\raisebox{0.5ex}{\footnotesize Scottish Gaelic (\texttt{gd-arcosg})}} \\
\hspace{1em}UDPipe 2  & 96.62 & 92.24 & 94.02 & 97.59 & 87.33 & 81.65 & 69.25 & 75.23\\
\hspace{1em}Stanza & 75.23 & 70.87 & 73.41 & 75.59 & 66.97 & 61.71 & 48.75 & 56.79 \\
\hspace{1em}BPE        & 97.66$^{\pm0.06}$ & 94.96$^{\pm0.07}$ & 95.82$^{\pm0.12}$ & 97.60$^{\pm0.04}$ & \uuline{\textbf{89.01}$^{\pm0.08}$} & \uuline{\textbf{85.63}$^{\pm0.15}$} & 75.81$^{\pm0.18}$ & 79.14$^{\pm0.18}$\\ 
\hspace{1em}\textsc{Factorizer} & \uuline{\textbf{97.78}$^{\pm0.05}$} & \uuline{\textbf{95.78}$^{\pm0.08}$} & \uuline{\textbf{96.67}$^{\pm0.09}$} & \uuline{\textbf{98.09}$^{\pm0.04}$} & 88.84$^{\pm0.16}$ & 85.30$^{\pm0.15}$ & \uuline{\textbf{77.00}$^{\pm0.22}$} & \uuline{\textbf{79.85}$^{\pm0.13}$} \\ \midrule

\multicolumn{5}{@{}l}{\raisebox{0.5ex}{\footnotesize Turkish (\texttt{tr-kenet})}} \\
\hspace{1em}UDPipe 2  & 93.72 & — & 92.06 & \textbf{93.33} & 84.07 & 71.29 & 61.92 & 64.89\\
\hspace{1em}Stanza & 92.98 & --- & 90.85 & 93.31 & \textbf{89.00} & \textbf{76.94} & 63.11 & 67.02  \\
\hspace{1em}BPE                    & 
\textbf{94.02}$^{\pm0.10}$ & --- & \textbf{92.22}$^{\pm0.04}$ & 92.60$^{\pm0.10}$ & \uuline{86.35$^{\pm0.07}$} & 74.60$^{\pm0.21}$ & 65.17$^{\pm0.30}$ & 67.71$^{\pm0.25}$ \\
\hspace{1em}\textsc{Factorizer}            & 93.94$^{\pm0.09}$ & --- & \textbf{92.22}$^{\pm0.11}$ &  \uuline{92.95$^{\pm0.08}$} & 86.18$^{\pm0.04}$ & 74.49$^{\pm0.13}$ & \textbf{65.26}$^{\pm0.14}$ & \uuline{\textbf{68.03}$^{\pm0.05}$} \\ \midrule

\multicolumn{5}{@{}l}{\raisebox{0.5ex}{\footnotesize Macro average}} \\
\hspace{1em}UDPipe 2  & 97.00 & 95.89 & 96.23 & 97.48 & 89.90 & 85.44 & 78.10 & 79.53 \\
\hspace{1em}Stanza & 88.54 & 82.67 & 88.45 & 87.56 & 79.51 & 75.20 & 67.03 & 70.93  \\
\hspace{1em}BPE                    & 
\textbf{97.50}$^{\pm0.06}$ & 96.79$^{\pm0.06}$ & 96.89$^{\pm0.06}$ & 97.34$^{\pm0.05}$ & 91.29$^{\pm0.09}$ & 87.44$^{\pm0.14}$ & 80.98$^{\pm0.19}$ & 83.00$^{\pm0.15}$
\\ 
\hspace{1em}\textsc{Factorizer}            & 
97.48$^{\pm0.06}$ & \textbf{96.94}$^{\pm0.06}$ & \textbf{97.07}$^{\pm0.06}$ & \textbf{97.67}$^{\pm0.04}$ & \textbf{91.45}$^{\pm0.10}$ & \textbf{87.60}$^{\pm0.13}$ & \textbf{81.46}$^{\pm0.16}$ & \textbf{83.65}$^{\pm0.12}$ \\ \bottomrule
\end{tabular}%
}
\caption{The results of finetuning our language models on treebanks from the Universal Dependencies project -- as compared to two well-performing publicly available parsers, UDPipe 2 \citep
{straka-etal-2019-udpipe} and Stanza \citep
{qi2020stanza}. Every model is evaluated on the test split of the respective treebank, we use the official evaluation script from the CoNLL 2018 shared task \citep[\footnotesize{\url{https://universaldependencies.org/conll18/evaluation.html}}]{zeman-etal-2018-conll} and report the mean and standard deviation over 5 runs with different random seeds, using the gold tokenization. The best results for each language are displayed in \textbf{bold}. In order to compare models, we use the Almost Stochastic Order test \citep[ASO;][]{del2018optimal, dror2019deep} implemented by \citet{ulmer2022deep}. We compare the BPE and \textsc{factorizer} models based on five random seeds using ASO with a confidence level of $\alpha = 0.05$ (before adjusting for comparison of 7 different languages using the Bonferroni correction). Almost stochastic dominance ($\epsilon_\text{min} < \tau$ with $\tau = 0.2$) is indicated by \uuline{underlines}.
}
\label{tab:multilingual}
\end{table*}
}

\subsection{Experiment 4: Robustness}

\paragraph{Robustness to noise.} Being able to maintain performance when dealing with unnormalized and noisy text, is an important quality of any NLP tool deployed in real-world scenarios. In order to evaluate how \textsc{factorizer} deals with unexpected noise, we finetune a UD parser on clean data using the \texttt{en-ewt} corpus and evaluate it on modified development sets with increasing levels of character noise. The noise is introduced by perturbing every character with a set probability $p_{\textrm{noise}}$ by uniformly choosing from (1) deleting the character, (2) changing its letter case, or (3) repeating the character 1--3 times.

\begin{figure}[!t]
    \includegraphics[width=0.995\columnwidth]{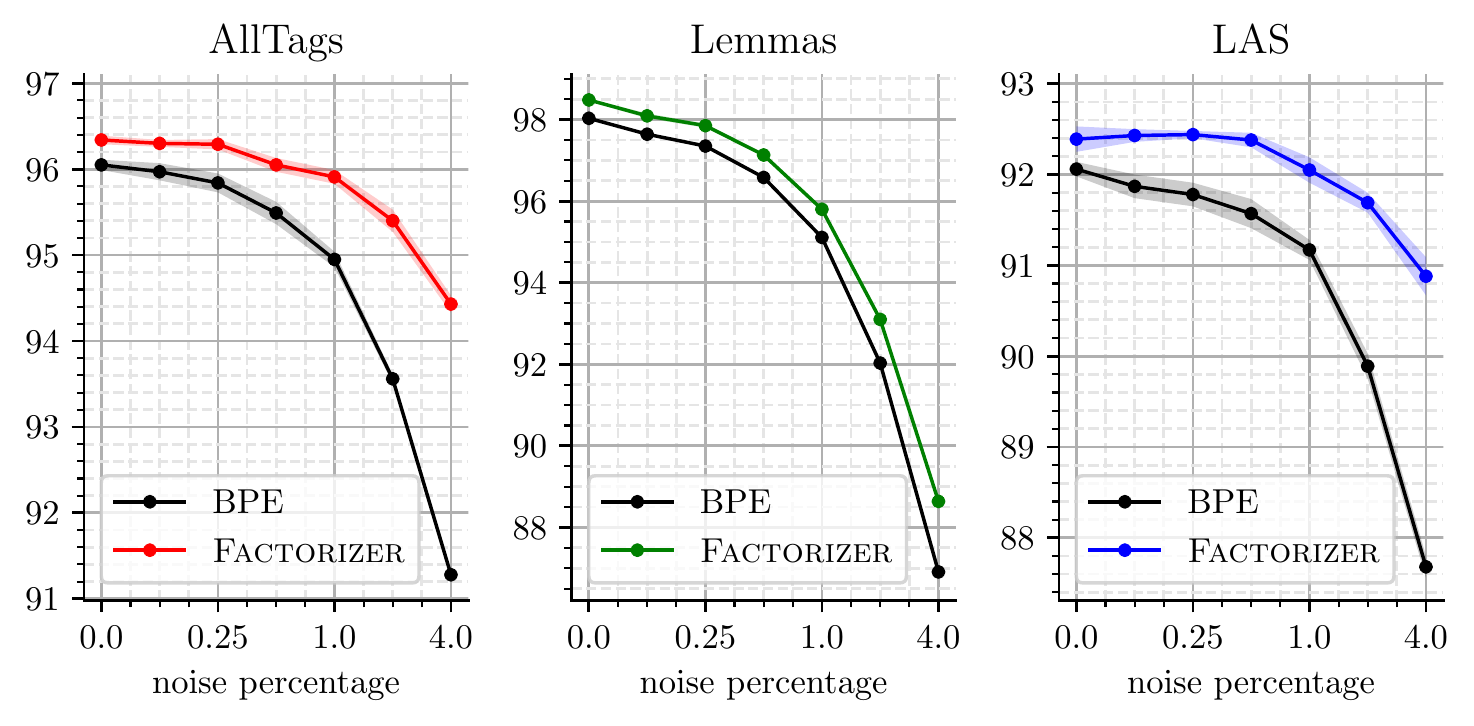}
    \caption{Robustness to noise. We train a single English parser on clean data and evaluate it on modified development sets with increasing amount of character noise. The performance of the \textsc{factorizer}-based parser is deteriorating slower than the BPE-based model.}
    \label{fig:noise}
\end{figure}

\cref{fig:noise} shows the relation between the noise level $p_{\textrm{noise}}$ and the expected performance. When comparing \textsc{factorizer} with BPE, it is apparent that \textsc{factorizer} is more robust to increased noise on tagging and dependency parsing as the performance drops more slowly with increased noise. The accuracy drops equally on lemmatization, which may be attributed to its formulation as prediction of relative lemma rules (as in UDPipe~2 by \newcite{straka-etal-2019-udpipe}).

\paragraph{Robustness to resource scarcity.} The promising results achieved on the very low-resource language of Scottish Gaelic have motivated a more thorough examination of the relationship between the size of a pretraining corpus and the downstream performance. In order to investigate this relationship, we pretrain multiple English language models using random fractions of the full English corpus while maintaining a constant number of training steps. 

\begin{figure}[!t]
    \includegraphics[width=\columnwidth]{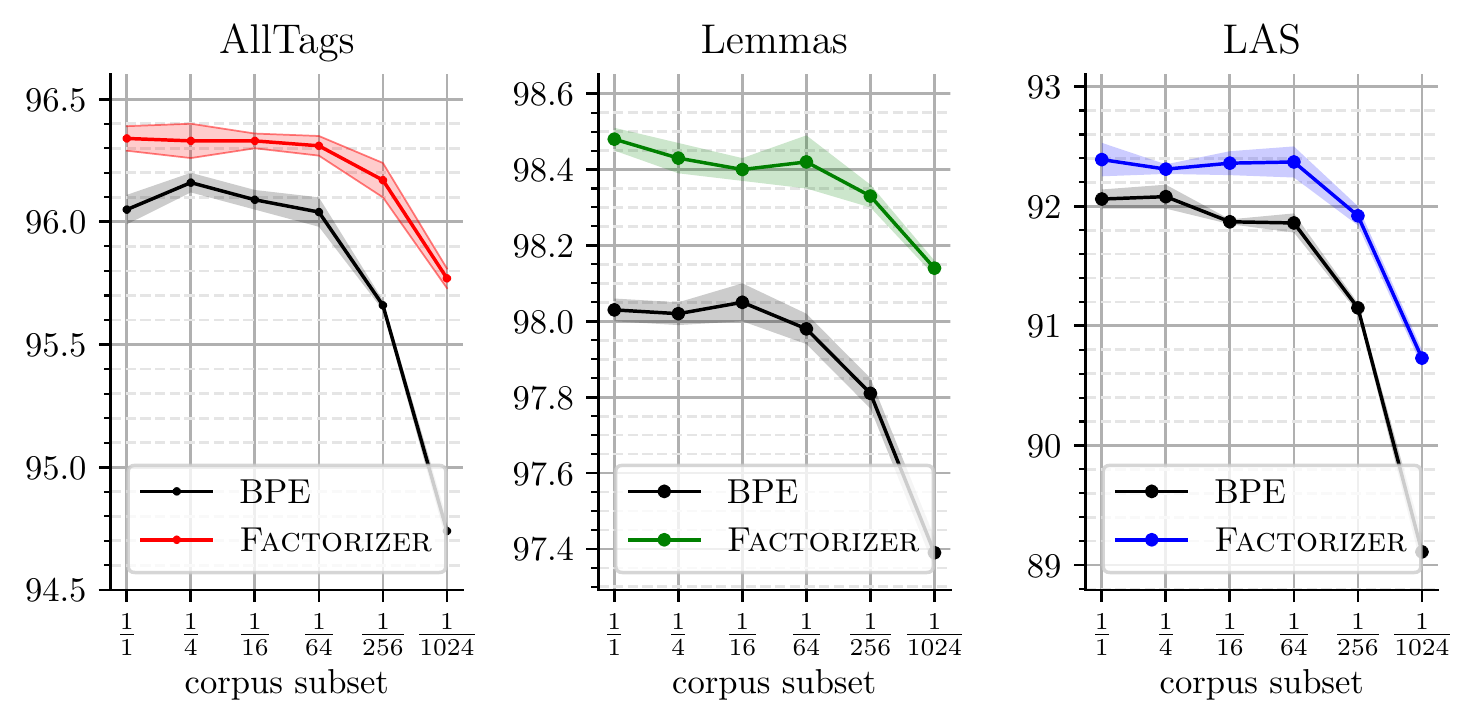}
    \caption{Robustness to data scarcity during pretraining. We pretrain language models on random fractions of the English corpus and then measure their performance when finetuned on \texttt{en-ewt}. We can see that \textsc{factorizer}-based language models are more robust to pretraining in low-resource settings.}
    \label{fig:subset}
\end{figure}

The performance of the finetuned models is evaluated and plotted in \cref{fig:subset}. The results demonstrate that our proposed method is more robust to resource scarcity. All language models are able to maintain performance with \nicefrac{1}{64} of the full corpus and after this point, the \textsc{factorizer}-based models are less sensitive to reducing the corpus size even more.

\section{Related work}

The neural era in NLP has brought about a change in how sentences are typically \emph{tokenized} into atomic units. Tokenization in current NLP typically involves the segmentation of the sentence into subwords, smaller units than the traditional word tokens of the pre-neural era. The question of what these subword units should consist of has received a fair amount of attention in previous research. Linguistically motivated or rule-based approaches are found \citep{porter1980algorithm}, however the majority of this work is based on unsupervised morphological segmentation. While this field has a long-standing history \citep{mielke2021between}, subwords have in recent years been based on BPE \citep{10.5555/177910.177914, sennrich-etal-2016-neural} and variations such as WordPiece \citep{https://doi.org/10.48550/arxiv.1609.08144} or SentencePiece \citep{kudo-richardson-2018-sentencepiece}.

There have been quite a few studies examining the influence of different morphological systems on language modeling performance \citep{vania-lopez-2017-characters, gerz-etal-2018-relation, bostrom-durrett-2020-byte}. In a highly multilingual setting, examining 92 languages, \newcite{10.1162/tacl_a_00365} study the influence of morphology on language modeling difficulty, contrasting BPE with the Morfessor system \citep{10.1145/1187415.1187418} and a rule-based morphological segmenter. There has also been some work comparing BPE with alternative tokenizers for downstream applications, mostly within machine translation and largely with negative results \citep{ataman-federico-2018-evaluation, https://doi.org/10.48550/arxiv.1812.08621, 10.1007/978-3-030-00794-2_30}. In this work we instead examine several morpho-syntactic tasks, finding clear improvements over BPE.

The recent pixel-based encoder of language \cite[PIXEL;][]{rust-etal-2023-pixel} reframes language modeling as a visual recognition task, employing a transformer-based encoder-decoder trained to reconstruct the pixels in masked image patches, and disposing of the vocabulary embedding layer completely. The method shows strong results for unseen scripts, however, more modest performance for languages with Latin script, such as English. As in our work, they find morpho-syntactic tasks to benefit the most and more semantic tasks to show more mixed results.


\section{Conclusion}

We proposed a novel tokenization method where every subword token is factorized into triplets of indices. The main benefit of this factorization is that the subwords-units maintain some character-level information without increasing the length of tokenized sequences. In practice, \textsc{factorizer} even slightly decreases the number of subword splits (\cref{fig:alpha-splits}), while noticeably improving the performance of large language models on morpho-syntactic tasks, especially on lemmatization (\cref{tab:multilingual}). Further experiments demonstrated increased robustness to noise and to data scarcity (\cref{fig:noise} and \ref{fig:subset}).

We hope that this work clearly demonstrated the potential of utilizing factorized subwords for language modeling and future work will improve their performance even further.


\section*{Limitations}

\paragraph{Size.} A limiting factor for some low-resource applications might be the size of a saved \textsc{factorizer} file. We only have to store the subword vocabulary, which however takes substantially more space than BPE as it needs to be stored as DAWG trie to keep the tokenization speed similar to BPE. For example, the saved English \textsc{factorizer} takes about 115MB of space while the English BPE with 32K subwords takes just about 1MB of space. We believe that the space requirements are negligable compared to the size of large language models, but they can a limiting factor in some edge cases.

The parameter-efficiency of \textsc{factorizer} follows the basic nature of factorized representations: a sequence of 3 bytes can represent more than 16M values ($256^3$). That’s how we can embed millions of subwords with a negligible parameter count. On the other hand, when we store a vocabulary with millions of subwords on disc, it necessarily requires more space than a BPE vocabulary with 10\,000s of subwords.

\paragraph{GLUE performance.} While our main interest and focus has been on morpho-syntactic downstream tasks, it is reasonable to ask what is the performance of \textsc{factorizer}-based language models on other tasks, such as natural language understanding. We utilize the pretrained English language models and finetune them on 8 GLUE tasks \citep{wang-etal-2018-glue}. The results in \cref{sec:glue} indicate that in this setting, our method is comparable to BPE but not better on average, even though both approaches stay within a standard deviation from each other. We hope that these results can be improved in future work.

\section*{Acknowledgements}
We would like to thank Petter Mæhlum, Andrey Kutuzov and Erik Velldal for providing very useful feedback on this work.

The efforts described in the current paper were jointly funded by the HPLT project (High Performance Language Technologies; coordinated by Charles University).

The computations were performed on resources provided through Sigma2 -- the national research infrastructure provider for High-Performance Computing and large-scale data storage in Norway.

\bibliography{anthology,custom}
\bibliographystyle{acl_natbib}

\clearpage
\onecolumn
\appendix

\section{Color interpretation}
\label{sec:color}

Our subword encoding method factorizes every subword-unit into $3\times256$ values, which is reminiscent of the standard 24-bit RGB color encoding. Thus every subword can be directly visualized as one pixel with an RGB color. While this interpretation does not have any practical application, it might help to explain our method from a different angle. Note that this analogy is not perfect -- our method views all indices as completely independent and thus the factorized `R channel' with index $101$ is as close to $102$ as to $237$.

\begin{figure*}[h]
    \includegraphics[trim={-1.25em 0 0 0}, width=0.9\textwidth]{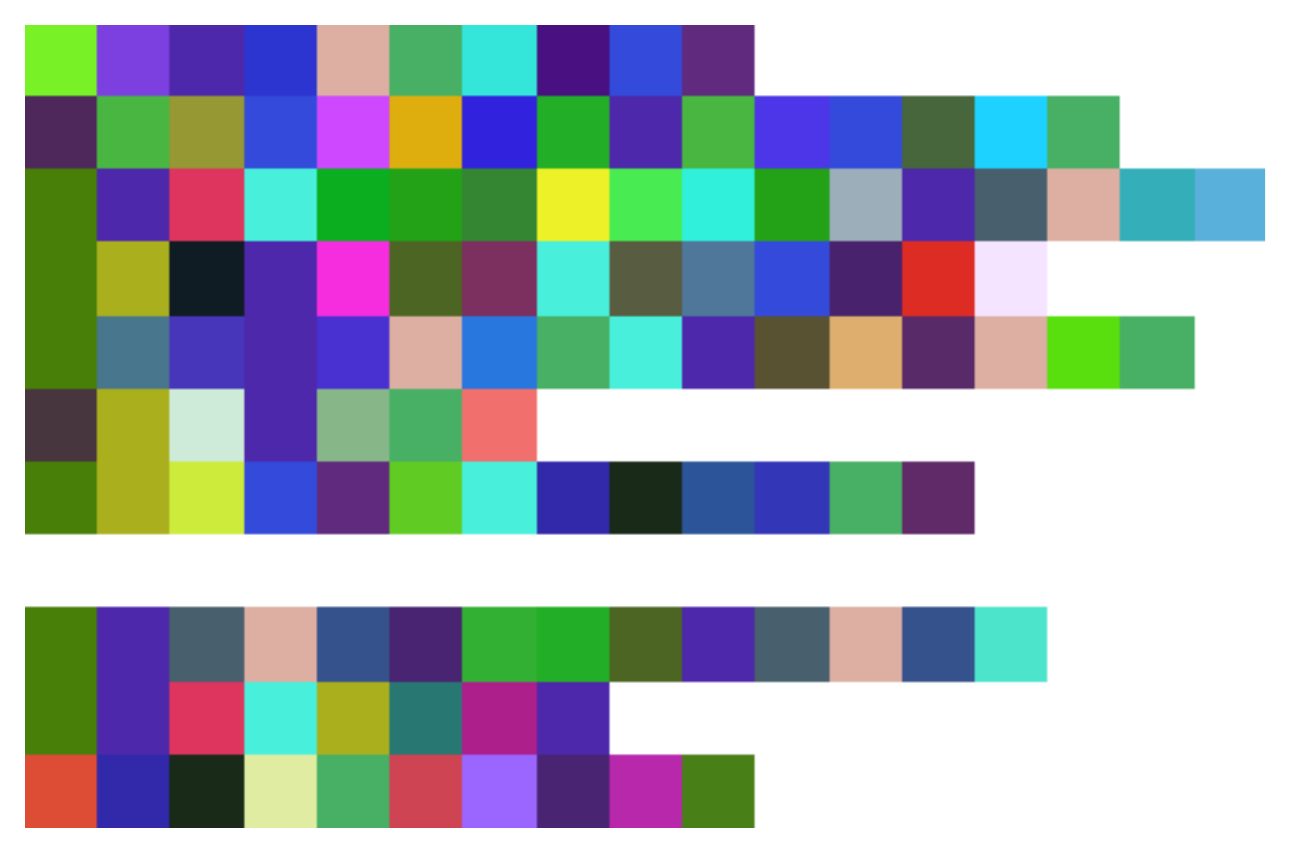}
    \caption{The first verse and chorus of the song Suzanne by Leonard Cohen, encoded with \textsc{factorizer}. Every subword can be interpreted as one 24-bit RGB pixel because we factorize every subword into $3\times256$ values. Notice how the colors of \textit{`river'} $[96, 42, 127]$, \textit{`lover`} $[96, 42, 104]$ and \textit{`love'} $[88, 42, 104]$ appear similar because these (sub)words share two out of three indices.}
    \label{fig:suzanne}
\end{figure*}

We illustrate this interpretation in \cref{fig:suzanne}, where we show the factorized subwords of the first verse and chorus of the song Suzanne by Leonard Cohen:

\textit{
\begin{verse}
Suzanne takes you down to her place near the river\\
You can hear the boats go by, you can spend the night beside her\\
And you know that she's half-crazy but that's why you want to be there\\
And she feeds you tea and oranges that come all the way from China\\
And just when you mean to tell her that you have no love to give her\\
Then she gets you on her wavelength\\
And she lets the river answer that you've always been her lover\\
~\\
And you want to travel with her, and you want to travel blind\\
And you know that she will trust you\\
For you've touched her perfect body with your mind
\end{verse}
}

\newpage
\section{Full English ablation results}
\label{sec:ablation-full}

The full results of the comparative study from \cref{sec:ablation} (using the full set of UD metrics) are given in \cref{tab:full-comparison}. This table also shows what happens if we keep a constant number of layers between the \textsc{factorizer} and BPE-based language models -- as opposed to fixing the parameter count.

\vspace{0.5em}

{\renewcommand{\arraystretch}{1.33}
\begin{table*}[h]
\resizebox{\textwidth}{!}{%
\begin{tabular}{@{}l@{\hspace{2em}}l@{\hspace{2em}}llllllll@{}}
\toprule
\textbf{Model}     & \textbf{\# params} & \textbf{UPOS}     & \textbf{XPOS}      & \textbf{UFeats}   & \textbf{Lemmas}   & \textbf{UAS}      & \textbf{LAS}      & \textbf{MLAS}     & \textbf{BLEX}     \\ \midrule

BPE  & 110.8M & 97.74$^{\pm0.02}$ & 97.55$^{\pm0.03}$ & 97.72$^{\pm0.04}$ & 98.03$^{\pm0.03}$ & 93.79$^{\pm0.10}$ & 92.06$^{\pm0.08}$ & 86.22$^{\pm0.11}$ & 87.94$^{\pm0.09}$\\
BPE + sampling  & 110.8M & 97.88$^{\pm0.02}$ & 97.71$^{\pm0.05}$ & 97.82$^{\pm0.05}$ & 98.25$^{\pm0.01}$ & 93.94$^{\pm0.07}$ & 92.22$^{\pm0.05}$ & 86.44$^{\pm0.04}$ & 88.35$^{\pm0.06}$ \\
BPE + sampling + averaging              & 110.8M & 97.88$^{\pm0.04}$ & 97.72$^{\pm0.02}$ & 97.81$^{\pm0.05}$ & 98.29$^{\pm0.04}$ & 93.89$^{\pm0.13}$ & 92.22$^{\pm0.12}$ & 86.58$^{\pm0.09}$ & 88.47$^{\pm0.09}$ \\[0.5em]
\textsc{Factorizer} & 108.1M             & 
97.76$^{\pm0.04}$ & 97.58$^{\pm0.04}$ & 97.74$^{\pm0.03}$ & 98.01$^{\pm0.01}$ & 94.04$^{\pm0.13}$ & 92.25$^{\pm0.13}$ & 86.26$^{\pm0.14}$ & 88.02$^{\pm0.20}$ \\ 
\textsc{Factorizer} + sampling & 108.1M & 97.94$^{\pm0.02}$ & 97.75$^{\pm0.06}$ & 97.84$^{\pm0.06}$ & 98.40$^{\pm0.02}$ & 93.96$^{\pm0.06}$ & 92.29$^{\pm0.07}$ & 86.62$^{\pm0.11}$ & 88.71$^{\pm0.12}$ \\
\textsc{Factorizer} + sampling + averaging & 108.1M & \textbf{97.97}$^{\pm0.02}$ & \textbf{97.77}$^{\pm0.03}$ & \textbf{97.87}$^{\pm0.02}$ & \textbf{98.48}$^{\pm0.03}$ & \textbf{94.06}$^{\pm0.13}$ & \textbf{92.39}$^{\pm0.14}$ & \textbf{86.87}$^{\pm0.19}$ & \textbf{88.92}$^{\pm0.19}$ \\[0.5em]
\textsc{Factorizer} (12L) & 86.8M             & 
97.75$^{\pm0.02}$ & 97.54$^{\pm0.03}$ & 97.74$^{\pm0.03}$ & 98.01$^{\pm0.05}$ & 93.72$^{\pm0.08}$ & 91.93$^{\pm0.08}$ & 85.99$^{\pm0.19}$ & 87.69$^{\pm0.20}$ \\
\textsc{Factorizer} (12L) + sampling & 86.8M             & 97.80$^{\pm0.03}$ & 97.68$^{\pm0.04}$ & 97.72$^{\pm0.07}$ & 98.24$^{\pm0.03}$ & 93.45$^{\pm0.11}$ & 91.66$^{\pm0.08}$ & 85.46$^{\pm0.09}$ & 87.61$^{\pm0.07}$ \\
\textsc{Factorizer} (12L) + sampling + averaging & 86.8M            & 97.88$^{\pm0.04}$ & 97.70$^{\pm0.04}$ & 97.76$^{\pm0.04}$ &  98.31$^{\pm0.03}$ & 93.75$^{\pm0.06}$ & 92.00$^{\pm0.09}$ & 86.06$^{\pm0.20}$ & 88.19$^{\pm0.10}$\\ \bottomrule
\end{tabular}%
}
\caption{Full comparison of BPE and \textsc{factorizer} configurations. `\textsc{Factorizer} (12L)' denote the configuration without any additional transformer layers. The metrics are measured on the development split of \texttt{en-ewt}. We show the mean and standard deviation statistics over 5 random seeds, the best results are displayed in bold.}
\label{tab:full-comparison}
\end{table*}
}

\section{GLUE results}
\label{sec:glue}

As noted in the Limitations section, our method is comparable to BPE but not better on average, when evaluated on this NLU benchmark suite. The detailed results are given in \cref{tab:finegrained-glue}.

\vspace{0.5em}

{\renewcommand{\arraystretch}{1.33}
\begin{table*}[h]
\centering
\resizebox{\textwidth}{!}{%
\begin{tabular}{@{}ll@{\hspace{2em}}lll@{\hspace{3em}}lll@{}}
\toprule
\multirow{3}{*}{\textbf{Task}} & \multirow{3}{*}{\textbf{Metric}} & \textbf{BPE} & \textbf{BPE} & \textbf{BPE} & \textbf{\textsc{Factorizer}} & \textbf{\textsc{Factorizer}} & \textbf{\textsc{Factorizer}} \\
 & & & \textbf{+ sampling} & \textbf{+ sampling} & & \textbf{+ sampling} & \textbf{+ sampling} \\
 & & & & \textbf{+ averaging} & & & \textbf{+ averaging} \\ \midrule
CoLA                           & MCC   & 59.56$^{\pm1.33}$ & 59.93$^{\pm1.74}$ & 60.25$^{\pm1.71}$ & \textbf{62.02}$^{\pm1.82}$ & 60.71$^{\pm1.29}$ & 60.42$^{\pm2.28}$        \\ [0.5em]
\multirow{1}{*}{MNLI}          & matched acc.             & \textbf{87.0}8$^{\pm0.29}$ & 86.03$^{\pm0.15}$ & 86.18$^{\pm0.07}$ & 86.35$^{\pm0.16}$ & 85.60$^{\pm0.07}$ & 85.86$^{\pm0.16}$ \\
                               & mismatched acc.   & 86.48$^{\pm0.20}$ & 86.46$^{\pm0.22}$ & \textbf{86.49}$^{\pm0.15}$ & 86.26$^{\pm0.11}$ & 85.61$^{\pm0.15}$ & 85.77$^{\pm0.23}$ \\ [0.5em]
\multirow{1}{*}{MRPC}          & accuracy                    & \textbf{89.61}$^{\pm0.45}$ & 88.77$^{\pm0.72}$ & 88.48$^{\pm0.71}$ & 86.96$^{\pm1.18}$ & 86.96$^{\pm0.40}$ & 87.79$^{\pm1.32}$ \\
                               & F\textsubscript{1} & \textbf{92.60}$^{\pm0.45}$ & 92.00$^{\pm0.72}$ & 91.79$^{\pm0.71}$ & 90.60$^{\pm1.18}$ & 90.74$^{\pm0.40}$ & 91.27$^{\pm1.32}$ \\ [0.5em]
QNLI                           & accuracy            & 91.98$^{\pm0.22}$ & 91.93$^{\pm0.12}$ & 91.93$^{\pm0.10}$ & \textbf{92.01}$^{\pm0.20}$ & 91.14$^{\pm0.22}$ & 91.58$^{\pm0.36}$ \\ [0.5em]
\multirow{1}{*}{QPP}           & accuracy             & 91.59$^{\pm0.05}$ & 91.62$^{\pm0.11}$ & 91.63$^{\pm0.10}$ & \textbf{91.75}$^{\pm0.14}$ & 91.69$^{\pm0.05}$ & 91.72$^{\pm0.07}$ \\
                               & F\textsubscript{1} & 88.78$^{\pm0.08}$ & 88.78$^{\pm0.16}$ & 88.86$^{\pm0.13}$ & \textbf{89.04}$^{\pm0.18}$ & 88.92$^{\pm0.08}$ & 88.94$^{\pm0.09}$ \\ [0.5em]
                               RTE                            & accuracy               & 73.14$^{\pm2.42}$ & 73.94$^{\pm1.09}$ & \textbf{75.31}$^{\pm1.16}$ & 66.35$^{\pm1.41}$ & 68.88$^{\pm1.23}$ & 71.48$^{\pm1.37}$         \\ [0.5em]
SST-2                          & accuracy    & 93.60$^{\pm0.05}$ & 93.62$^{\pm0.51}$ & 93.76$^{\pm0.48}$ & \textbf{93.83}$^{\pm0.41}$ & 93.07$^{\pm0.28}$ & 93.30$^{\pm0.43}$     \\ [0.5em]
\multirow{1}{*}{STS-B}         & Pearson corr.       & 89.21$^{\pm0.34}$ & 89.26$^{\pm0.42}$ & \textbf{89.67}$^{\pm0.46}$ & 89.32$^{\pm0.25}$ & 88.71$^{\pm0.10}$ & 88.84$^{\pm0.19}$ \\
                               & Spearman corr.    & 89.23$^{\pm0.27}$ & 89.14$^{\pm0.33}$ & \textbf{89.52}$^{\pm0.39}$ & 89.06$^{\pm0.20}$ & 88.48$^{\pm0.09}$ & 88.61$^{\pm0.20}$ \\ [0.5em] \midrule
\textbf{Average} &                           & 84.45$^{\pm1.00}$ & 84.43$^{\pm0.81}$ & \textbf{84.70}$^{\pm0.81}$ & 83.61$^{\pm0.93}$ & 83.40$^{\pm0.66}$ & 83.90$^{\pm1.07}$ \\ \bottomrule
\end{tabular}%
}
\caption{Detailed development GLUE results for the English language models. We show the mean and standard deviation statistics over 5 runs with different random seeds.}
\label{tab:finegrained-glue}
\end{table*}
}

\newpage
\section{Hyperparameters}
\label{sec:hyperparameters}

All hyperparameters are listed below: VQ-VAE hyperparameters in \cref{tab:vae-hyperparams}, pretraining hyperparameters in \cref{tab:lm-hyperparams} and finetuning hyperparameters in \cref{tab:ud-hyperparams}. Note that the the full PyTorch \citep{NEURIPS2019_9015} source code can be found in {\small\url{https://github.com/ltgoslo/factorizer}}.

The training was performed on 128 AMD MI250X GPUs (distributed over 16 compute nodes) and took approximately 8 hours per model in a mixed precision mode.

\vspace{5em}

\begin{table*}[h]
\centering
\begin{tabular}{@{}lcccc@{}}
\toprule
\textbf{Hyperparameter} & \textbf{Value} \\ \midrule
Codebook size & 256 \\
Number of codebooks & 3 \\
Number of layers        & 6          \\
Hidden size             & 256           \\
FF intermediate size    & 683        \\
FF activation function  & GEGLU         \\
Attention heads         & 4            \\
Attention head size     & 64            \\
Dropout                 & 0.1           \\
Attention dropout       & 0.1           \\
Training steps          & 50\,000       \\
Batch size              & 4\,096  \\
Warmup steps            & 500        \\
Initial learning rate   & 0.001          \\
Final learning rate     & 0.0001          \\
Learning rate decay     & cosine        \\
Weight decay            & 0.01          \\
Layer norm $\epsilon$   & 1e-5          \\
Optimizer               & ASAM + AdamW         \\
ASAM $\rho$         & 0.2          \\ 
AdamW $\epsilon$         & 1e-6          \\
AdamW $\beta_1$          & 0.9           \\
AdamW $\beta_2$          & 0.98          \\
VQ-VAE $\beta$ & 0.5 \\
Codebook EMA $\lambda$ & 0.96 \\
Weight EMA $\lambda$ & 0.999 \\
\bottomrule
\end{tabular}
\caption{VQ-VAE hyperparameters.}
\label{tab:vae-hyperparams}
\end{table*}

\begin{table*}[h]
\centering
\begin{tabular}{@{}lcccc@{}}
\toprule
\textbf{Hyperparameter} & \textbf{Value} \\ \midrule
Number of layers        & 12 or 15           \\
Hidden size             & 768           \\
FF intermediate size    & 2\,048        \\
Vocabulary size         & 32\,768 or $3 \times 256$         \\
FF activation function  & GEGLU         \\
Attention heads         & 12            \\
Attention head size     & 64            \\
Dropout                 & 0.1           \\
Attention dropout       & 0.1           \\
Training steps          & 31\,250       \\
Batch size              & 32\,768 (90\% steps) / 8\,192 (10\% steps)        \\
Sequence length         & 128 (90\% steps) / 512 (10\% steps)         \\
Tokens per step         & 4\,194\,304   \\
Warmup steps            & 500 (1.6\% steps)         \\
Initial learning rate   & 0.01          \\
Final learning rate     & 0.001          \\
Learning rate decay     & cosine        \\
Weight decay            & 0.1          \\
Layer norm $\epsilon$   & 1e-5          \\
Optimizer               & LAMB         \\
LAMB $\epsilon$         & 1e-6          \\
LAMB $\beta_1$          & 0.9           \\
LAMB $\beta_2$          & 0.98          \\
Gradient clipping       & 2.0           \\ 
\textsc{Factorizer} $\alpha_{\textrm{split}}$ (if applicable) & 0.1 \\
\textsc{Factorizer} $\sigma_{\textrm{sample}}$ (if applicable) & 0.0 or 0.02 \\
BPE dropout rate (if applicable) & 0.0 or 0.1 \\
\bottomrule
\end{tabular}
\caption{Language model pretraining hyperparameters.}
\label{tab:lm-hyperparams}
\end{table*}

\begin{table*}[h]
\centering
\begin{tabular}{@{}lcccc@{}}
\toprule
\textbf{Hyperparameter} & \textbf{Value} \\ \midrule
Hidden size             & 768           \\
Dropout                 & 0.2           \\
Attention dropout       & 0.2           \\
Word dropout & 0.15 \\
Label smoothing & 0.1 \\
Epochs & 20 \\
Batch size              & 32        \\
Warmup steps            & 250         \\
Initial learning rate   & 0.001          \\
Final learning rate     & 0.0001          \\
Learning rate decay     & cosine        \\
Weight decay            & 0.001          \\
Layer norm $\epsilon$   & 1e-5          \\
Optimizer               & AdamW         \\
AdamW $\epsilon$         & 1e-6          \\
AdamW $\beta_1$          & 0.9           \\
AdamW $\beta_2$          & 0.98          \\
Gradient clipping       & 10.0           \\ 
\textsc{Factorizer} $\alpha_{\textrm{split}}$ (if applicable) & 0.1 \\
\textsc{Factorizer} $\sigma_{\textrm{sample}}$ (if applicable) & 0.0 or 0.02 \\
BPE dropout rate (if applicable) & 0.0 or 0.1 \\
Samples for averaging (if applicable) & 128 \\
\bottomrule
\end{tabular}
\caption{Hyperparameters for finetuning language models on UD treebanks.}
\label{tab:ud-hyperparams}
\end{table*}

\end{document}